\documentclass{ecai} 

\usepackage[table]{xcolor}
\usepackage{colortbl}
\usepackage{latexsym}
\usepackage{amssymb}
\usepackage{amsmath}
\usepackage{amsthm}
\usepackage{booktabs}
\usepackage{enumitem}
\usepackage{graphicx}
\usepackage{color}
\usepackage{pifont}
\usepackage{amsfonts}
\usepackage{algorithm}
\usepackage{algorithmic}
\usepackage{array}
\usepackage{multirow}
\usepackage{arydshln}
\usepackage{longtable}
\usepackage[normalem]{ulem}
\useunder{\uline}{\ul}{}
\definecolor{lightgray}{gray}{0.9}
\usepackage{dsfont}

\usepackage{listings}
\newcommand{\BibTeX}{B\kern-.05em{\sc i\kern-.025em b}\kern-.08em\TeX}

\begin{document}


\begin{frontmatter}


\paperid{5113} 


\title{Latent Knowledge Scalpel: Precise and Massive Knowledge Editing for Large Language Models}


\author[A,B]{\fnms{Xin}~\snm{Liu}}
\author[A,B]{\fnms{Qiyang}~\snm{Song}}
\author[A,B]{\fnms{Shaowen}~\snm{Xu}} 
\author[C]{\fnms{Kerou}~\snm{Zhou}} 
\author[D]{\fnms{Wenbo}~\snm{Jiang}} 
\author[A,B]{\fnms{Xiaoqi}~\snm{Jia}\thanks{Corresponding Author. Email: jiaxiaoqi@iie.ac.cn}}
\author[A,B]{\fnms{Weijuan}~\snm{Zhang}} 
\author[A,B]{\fnms{Heqing}~\snm{Huang}} 
\author[A,B]{\fnms{Yakai}~\snm{Li}} 

\address[A]{Institute of Information Engineering, Chinese Academy of Sciences}
\address[B]{School of Cyber Security, University of Chinese Academy of Sciences}
\address[C]{Tsinghua University}
\address[C]{University of Electronic Science and Technology of China}


\begin{abstract}
Large Language Models (LLMs) often retain inaccurate or outdated information from pre-training, leading to incorrect predictions or biased outputs during inference. While existing model editing methods can address this challenge, they struggle with editing large amounts of factual information simultaneously and may compromise the general capabilities of the models. In this paper, our empirical study demonstrates that it is feasible to edit the internal representations of LLMs and replace the entities in a manner similar to editing natural language inputs. Based on this insight, we introduce the Latent Knowledge Scalpel (LKS), an LLM editor that manipulates the latent knowledge of specific entities via a lightweight hypernetwork to enable precise and large-scale editing. Experiments conducted on Llama-2 and Mistral show even with the number of simultaneous edits reaching 10,000, LKS effectively performs knowledge editing while preserving the general abilities of the edited LLMs. Code is available at: \url{https://github.com/Linuxin-xxx/LKS}.
\end{abstract}

\end{frontmatter}

\section{Introduction}
The development of large language models (LLMs) has significantly advanced natural language processing (NLP) \citep{llmmeet2024}. However, challenges such as hallucinations \citep{Hallucination2024, hallucinationinevitable2024}, biases \citep{bias2024}, and outdated information \citep{Mindthegap2024} persist after pre-training. Therefore, it is essential to perform targeted updates to this incorrect or outdated information that arises during the deployment of LLMs.

Retraining or fine-tuning \citep{wei2022finetuned} can address this issue but requires substantial computational resources and time. Parameter-efficient fine-tuning (PEFT) methods \citep{peft} provide more efficient alternatives, though they may lead to overfitting and are limited in reliability \citep{easyedit, KE}. Another class of methods modifies the behavior of LLMs by adding contextual information to the prompts, including prompt engineering \citep{PEsurvey2024} and retrieval-augmented generation (RAG) \citep{rag}. However, these methods may fail due to misalignment between LLMs and prompts \citep{hernandez2024inspecting}. Moreover, they are constrained by prompt length, as they require ample context to be effective \citep{easyedit}.

\begin{figure}[t]
  \includegraphics[width=\columnwidth]{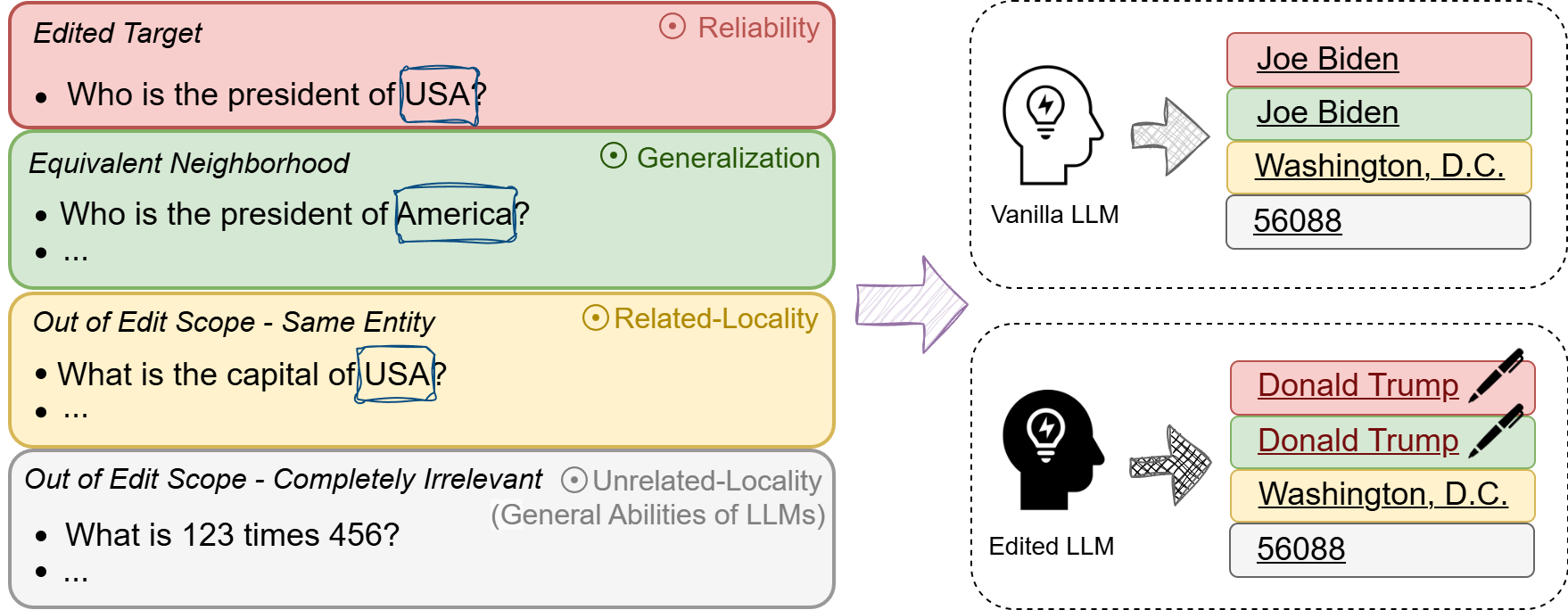}
  \caption{Illustration of model editing. Model editing modifies specific knowledge with minimal impact on unrelated inputs.}
  \label{fig:model-editing}
\end{figure}

Model editing has emerged as a promising solution \citep{kesurvey, editllmpmo}, aiming to make targeted modifications to specific model behaviors while minimizing changes to unrelated distributions, as shown in Figure~\ref{fig:model-editing}. While previous works have introduced various enlightening editing approaches, there remains room for improvement. \citet{harm2024} highlights that editing methods that modify model weights, such as \citet{KN}, \citet{MEND}, \citet{ROME}, and \citet{MEMIT}, can lead to overfitting on the edited facts, degrading the model’s general abilities. Furthermore, methods such as \citet{KE}, \citet{KN}, \citet{MEND}, and \citet{ROME} become less effective when editing large volumes of factual information simultaneously \citep{MEND, MEMIT}. \citet{GRACE} directly replaces the hidden states of the original model with the edit target to enable lifelong sequential editing, but it suffers from poor generalization and often fails to edit paraphrases of the targets. 

In this paper, we propose Latent Knowledge Scalpel (LKS), an LLM editor capable of performing large-scale simultaneous knowledge editing without compromising the general abilities of LLMs. Unlike methods that modify the model's weights, we focus on editing the internal representations of specific entities. Previous studies \citep{petroni2019, jiang2020, implicit2021, sun2024learninglearntesttime} have shown that the internal representations (or hidden states) of LLMs contain both factual knowledge and contextual information. For fine-grained editing, we associate knowledge with entities, which represent the smallest unit of knowledge in natural language \citep{Autoregressive21}. Our empirical study (\S \ref{sec:em_study}) demonstrates that the internal representation of a single entity encapsulates both factual knowledge and semantic features, which we refer to as a \textbf{knowledge block (KB)}. Moreover, we show that the internal representations of LLMs preserve the syntactic structure of natural language, allowing operations similar to those on natural language itself. 

Building on these findings, LKS manipulates specific entity latent knowledge for targeted updates (\S \ref{sec:method}). During inference, if the input contains an entity within the edit scope, LKS uses a simple neural network to generate a new knowledge block (KB) for this entity and replace the original one, guiding the LLM to produce the desired output. This network is trained to integrate the new knowledge of entities within the edit scope, enabling it to generate optimal KBs. These KBs update specific entity features while preserving others, ensuring precise edits. Moreover, the use of the neural network allows LKS to handle large-scale, simultaneous updates. Our entity recognition mechanism ensures accurate identification of the edit scope, preventing LKS from triggering on inputs outside the scope, thereby enabling extensive edits without affecting unrelated distributions.

We conduct extensive experiments to evaluate our LKS editor (\S \ref{sec:exps}). Our experimental results demonstrate that LKS outperforms six other methods in factual knowledge editing on Llama-2-7B and Mistral-7B, achieving the best balance in reliability, generality, and locality. Additionally, during large-scale simultaneous editing, LKS can accurately perform 10,000 edits simultaneously, achieving high edit performance while maintaining the general abilities of the LLMs.

We make the following key contributions: 
\begin{enumerate}  
\item We introduce Latent Knowledge Scalpel (LKS), an LLM editor that replaces entity knowledge blocks with new ones generated by a simple neural network, achieving targeted and large-scale LLM editing while preserving the general abilities of LLMs. 
\item We demonstrate that the entity knowledge blocks in LLMs contain semantic information, and the internal representations of LLMs retain the syntactic structure of natural language, allowing us to manipulate them like natural language. 
\item Our experiments show that even when the number of simultaneous edits reaches 10,000, LKS is still able to maintain the general abilities of the edited LLMs while outperforming other editors in terms of edit performance.
\end{enumerate}

\section{Empirical Study}
\label{sec:em_study}
\subsection{Semantic Information of a Single Entity Knowledge Block}
In natural language, an entity typically contains multiple factual knowledge. For example, a person entity may include information such as age, occupation, and hobbies. This raises the question: does a single entity knowledge block from a LLM also contain sufficient semantic information?

To investigate this, we design a probe to differentiate between factual knowledge learned by the LLM and counterfactual knowledge it has not encountered. Specifically, we extract 10,000 entities along with their factual and counterfactual attributes from the Counterfact dataset \citep{ROME}. The probe computes the cosine similarity between the entity KB and the internal representations of the last tokens from both factual and counterfactual knowledge, selecting the one with the higher similarity:

\begin{small}
\begin{equation}
\mathop{argmax}\limits_{knowledge \in \mathcal{K}} \, cosine\text{-}similarity(R_{entity}, R_{knowledge})
\label{eq:cossim}
\end{equation}
\end{small}

\noindent where $\mathcal{K}$ contains both factual and counterfactual knowledge and $R$ denotes internal representation. The probe's accuracy is defined as the proportion of correctly selected factual knowledge. Higher accuracy indicates that the entity KB is semantically closer to learned knowledge, suggesting it encodes meaningful semantic information.

\begin{figure}[t]
  \includegraphics[width=\linewidth]{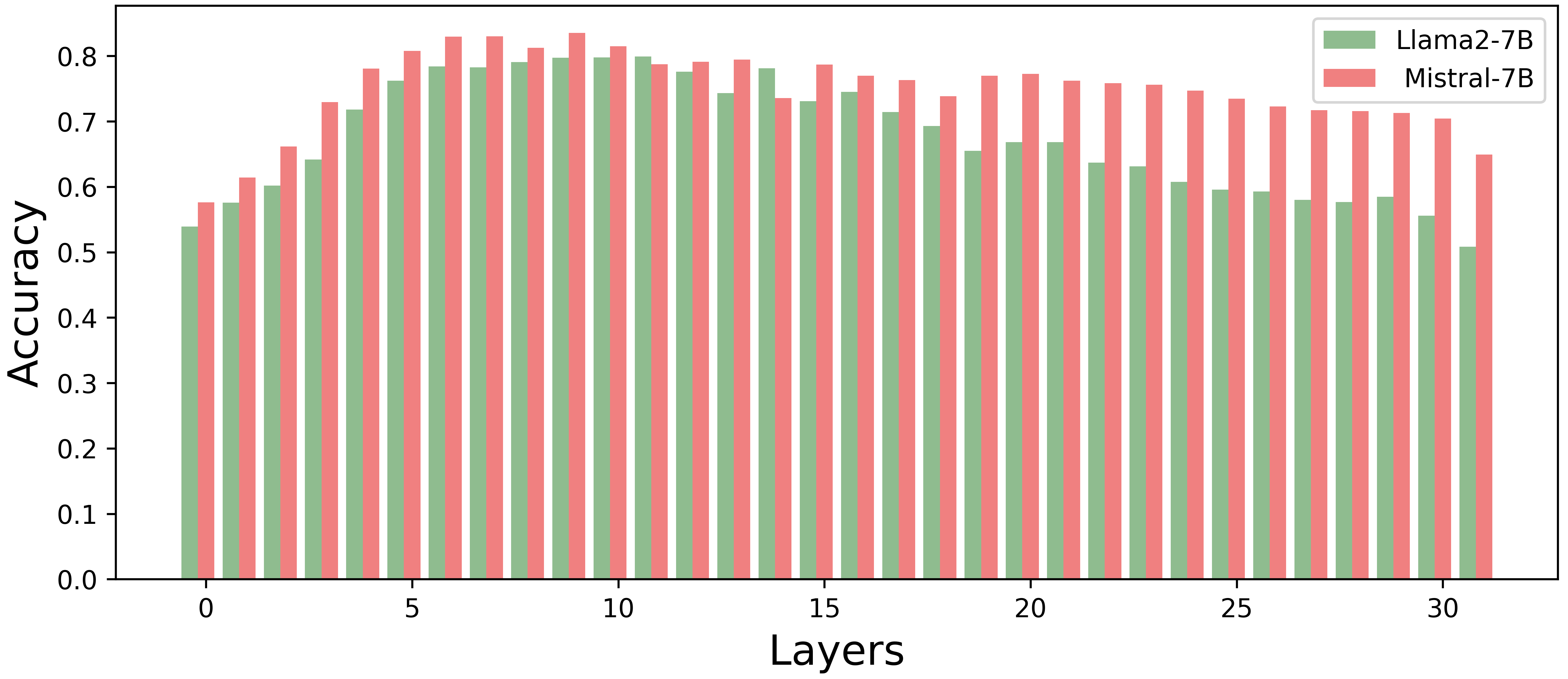}
  \caption{Probe accuracy for identifying factual knowledge across layers in Llama-2-7B and Mistral-7B. The results show that the probe accuracy exceeding 50\% on average and peaking at 80\%, demonstrating that a single entity KB retains semantic information. }
  \label{fig:siml-probe}
\end{figure}

Figure~\ref{fig:siml-probe} presents the probe's accuracy across layers in Llama-2-7B-Chat \citep{llama2} and Mistral-7B-Instruct-v0.3 \citep{mistral7b}. The probe achieves an average accuracy above 50\%, surpassing random guessing, with peak accuracy reaching 80\%. These results confirm that a single entity KB in a LLM retains its semantic information.

\subsection{Syntactic Structure of Internal Representations}
\begin{figure}[t]
  \includegraphics[width=\linewidth]{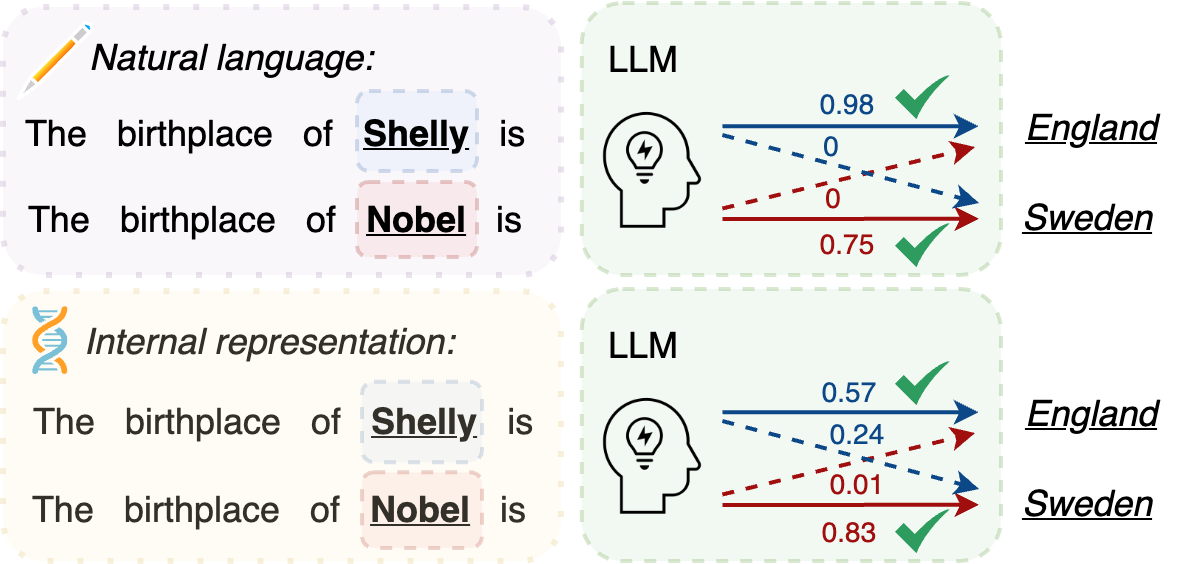}
  \caption{Upper: In natural language, replacing the entity "Shelly" with "Nobel" in the context of the "birthplace" causes the prediction from Llama-2-7B shifting from "England" to "Sweden". Lower: In internal representation, by obtaining the internal representations of two sentences and swapping the entity KB at a certain layer, similar to replacing entity names in a natural language prompt, the prediction of LLM changes and outputs the corresponding birthplaces.}
  \label{fig:syntactic}
\end{figure}

Natural language follows a syntactic structure, and replacing an entity name in a natural language prompt shifts the LLM’s prediction toward the semantics of the new entity. Our research shows that the internal representations of LLMs exhibit a similar syntactic structure, as illustrated in Figure~\ref{fig:syntactic}.

\begin{figure}[t]
  \includegraphics[width=\linewidth]{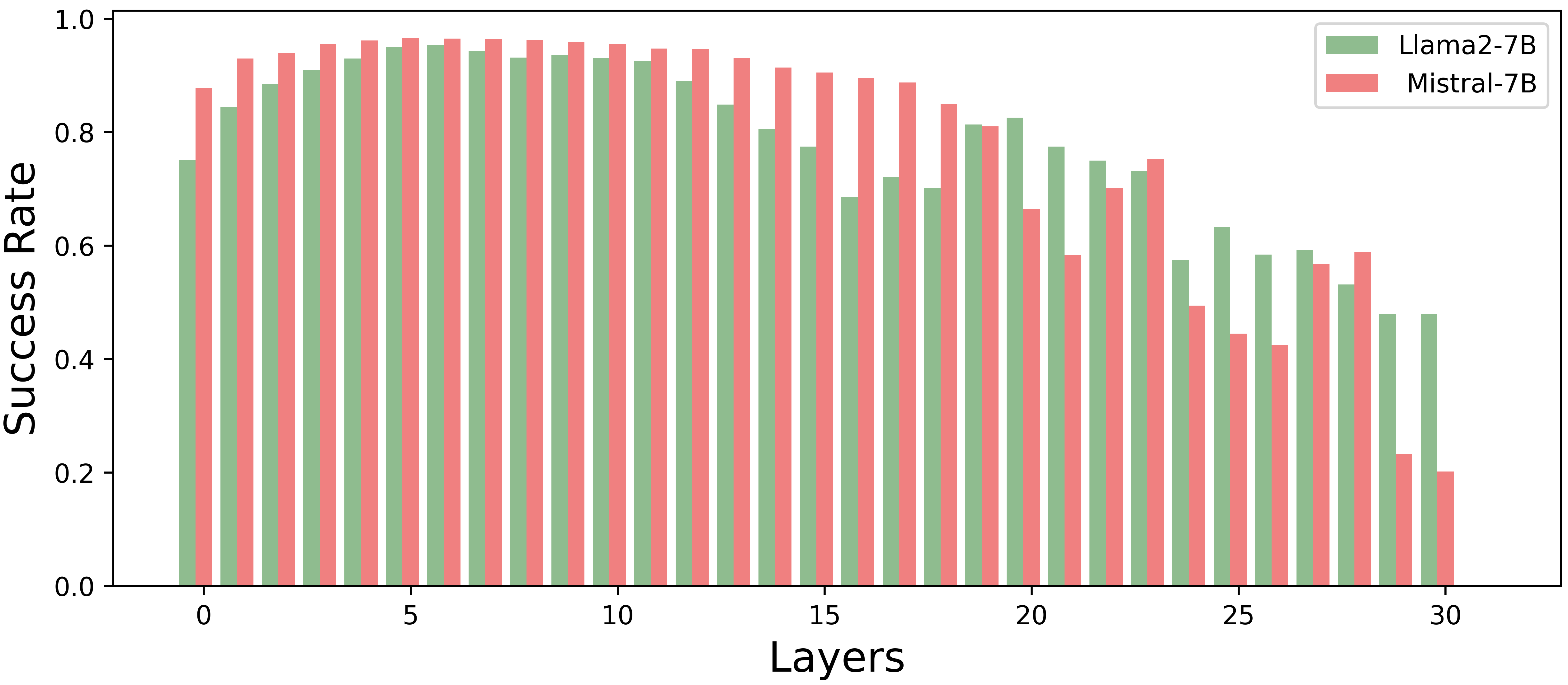}
  \caption{By replacing the name KB in the template with different entity KBs at each layer of Llama-2-7B and Mistral-7B, an increase in the ranking of the target birthplace across all layers in both models can be observed, confirming that internal representations of LLMs retain syntactic structure.}
  \label{fig:hs-cloze}
\end{figure}

To investigate this, we use the template "\texttt{The birthplace of Alfred Bernhard Nobel is}" and replace the KB of "\texttt{Alfred Bernhard Nobel}" with different entity KBs. We then measure the rate at which the predicted birthplaces rank higher after replacement. The results in Figure~\ref{fig:hs-cloze} show that replacing KBs increases the ranking of the target location across all layers in both Llama-2-7B and Mistral-7B. Additionally, the effect diminishes as the layer number increases.

These findings confirm that LLMs’ internal representations preserve syntactic structure to some extent. Furthermore, they suggest that during forward propagation, unchanged parts of the internal representation continue to influence predictions, explaining why the effect of KB replacement is stronger in earlier layers. If the goal is to introduce new information while preserving some original knowledge, modifying KBs in intermediate layers may be more effective.

\section{Method}
\label{sec:method}
\subsection{Overview of LKS}
\textbf{Design Goal} We aim to design an LLM editor that can effectively modify large-scale knowledge simultaneously while preserving the general abilities of LLMs. Particularly, it should satisfy the following requirements for LLM editing:
\begin{itemize}
\item Reliability: Accurately updates the specified targets.
\item Generality: Consistently updates the equivalent neighborhoods of the specified targets.
\item Locality: Ensures that knowledge outside the edit scope remains intact. 
\end{itemize}
 
We propose Latent Knowledge Scalpel (LKS), an LLM editor that precisely updates the latent knowledge of LLMs using a hypernetwork. We extract entity-related knowledge from an LLM, construct a self-supervised training dataset, and train a simple neural network (linear or MLP) specialized in entity-related knowledge. The new entity knowledge block (KB) generated by the network replaces the original one in the LLM. This updated entity KB is integrated into the LLM’s forward propagation, guiding the model to produce the edited target within the edit scope while preserving its original predictions outside this scope.

\begin{figure}[t]
  \includegraphics[width=\columnwidth]{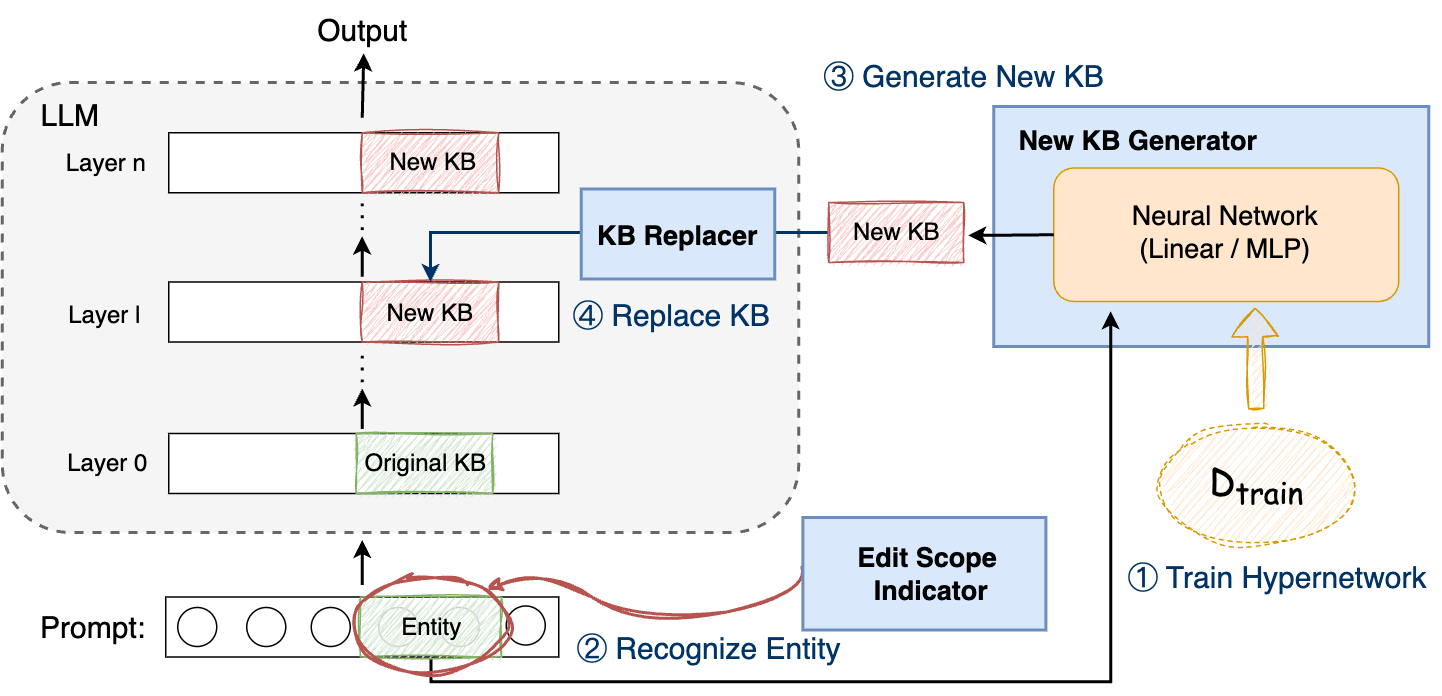}
  \caption{Architecture and Process of LKS. \ding{192} A simple neural network is trained using $\mathcal{D}_{train}$ to generate the optimal new KB during inference. \ding{193} Upon receiving a prompt, the Edit Scope Indicator checks if the target entity is present. If so, the relevant information is passed to the New KB Generator; otherwise, the original LLM proceeds as usual. \ding{194} The New KB Generator then creates the updated entity KB. \ding{195} The KB Replacer updates the corresponding entity KB in the selected layer $l$, and the inference continues to produce the final edited prediction.}
  \label{fig:KBRarch}
\end{figure}

The architecture of LKS is shown in Figure~\ref{fig:KBRarch}. LKS consists of three components: \textbf{Edit Scope Indicator}, which determines if an entity in the prompt falls within the edit scope, using fuzzy string matching and Levenshtein distance; \textbf{New KB Generator}, a simple neural network that generates the updated entity KB, which can either be a linear layer or an MLP layer. It is trained on a dataset containing the latest knowledge of entities within the edit scope, enabling it to output the optimal new entity KB; and \textbf{KB Replacer}, which hooks into a selected layer (discussed in detail in Section \ref{sec:selectlayer}) of the edited LLM and replaces the original entity KB with the new one generated by the New KB Generator. The updated entity KB is then involved in the LLM’s forward propagation, ultimately guiding the model’s prediction.

If the Edit Scope Indicator determines that the prompt contains the entity to be edited, the New KB Generator generates the updated entity KB for that entity. The KB Replacer then replaces the original entity KB in the selected layer, and the inference process continues until the edited LLM’s prediction is obtained. Otherwise, the last two components are not triggered, and the original model proceeds with the inference as usual.

\subsection{Building a New Knowledge Block} 
\begin{figure}[t]
  \includegraphics[width=\columnwidth]{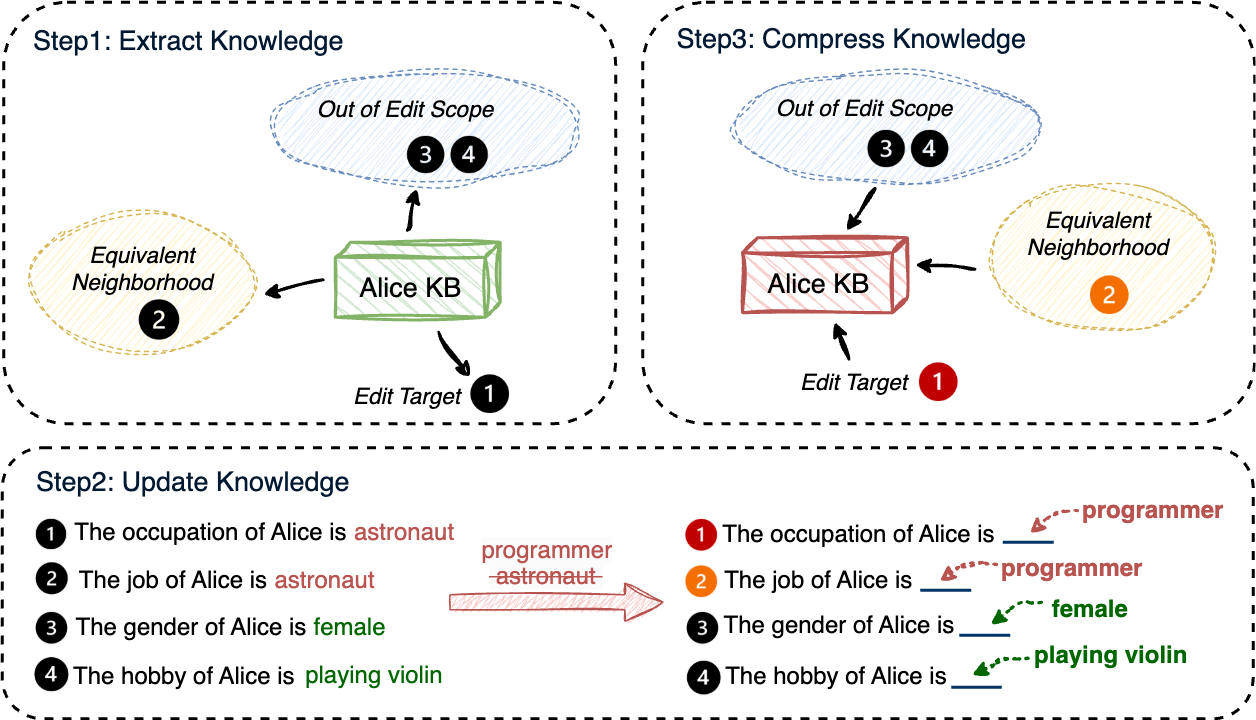}
  \caption{The process of building a new KB. \ding{192} Extract entity knowledge from a LLM. \ding{193} Update the target knowledge for editing the entity. \ding{194} Compress the knowledge using a simple neural network, which contains only the latest knowledge of entities within the edit scope.}
  \label{fig:buildKB}
\end{figure}

LKS enables LLMs to generate updated predictions for inputs within the edit scope (target edits and their equivalent neighborhoods) while preserving predictions outside this scope. In other words, it selectively edits a semantic feature of an entity while maintaining unrelated content. To achieve this, we construct a new knowledge block in three steps, as illustrated in Figure~\ref{fig:buildKB}.

\textbf{Knowledge Extraction} Inspired by \citet{commonsense}, we extract text-based entity-related knowledge from the LLMs. For each entity, we use GPT-4o mini \citep{gpt4} to generate multiple sentences reflecting its factual knowledge.

\textbf{Knowledge Updating} We replace the factual knowledge of the target feature and its equivalent neighborhood with the desired content, while leaving other entity features unchanged. These unchanged features will be aligned with the relevant knowledge in the edited LLM during the next step.

\textbf{Knowledge Compression} Following prior works \citep{petroni2019, autoprompt2020, roberts2020, entitycloze2022, position2022, metalearning2022, givefacts2023}, we convert the extracted and updated entity knowledge into gap-filling prompts to create a self-supervised training dataset $\mathcal{D}_{train}$. A simple neural network is then trained on $\mathcal{D}_{train}$, serving as a hypernetwork to generate new entity KBs that replace the original ones in the LLM. During training, the LLM aligns its predictions with the updated targets while retaining non-edited knowledge. After training, this neural network encapsulates only the latest entity knowledge and can produce the optimal new entity KBs which represent the compressed knowledge.

\subsection{Training LKS Hypernetwork} 
The neural network $h_{\phi}(\text{·})$ takes the input entity $E$ and outputs the new knowledge block for layer $l$, denoted as $\tilde{R}^l_{\phi}(E) = h_{\phi}(E;l)$. This hypernetwork is trained using $\mathcal{D}_{train}$ in advance to generate the optimal new KB $\tilde{R}^l$ during inference. During LLM inference, LKS replaces the original KB $R^l$ with the new KB $\tilde{R}^l$, guiding the LLM’s predictions. Notably, $\mathcal{D}_{train}$ is significantly smaller than the original LLM training dataset, and the storage overhead of the neural network is negligible compared to the LLM itself. For instance, $h_{\phi}$ with a linear layer for Llama-2-7B occupies only 64MB, regardless of the number of edits it contains.

Given an LLM $f_{\theta}$ and an input sequence $x$ containing entity $E$, the model recalls the corresponding feature of $E$ and predicts the token sequence $y$. The original entity KB in layer $l$ can be formulated as $R^l_{\theta}(E) = R^{l-1}_{\theta}(E) + attn^l_{\theta}(E) + mlp^l_{\theta}(E)$. The output $y$ can be expressed as $y = f_{\theta}(x, R^l_{\theta}(E))$. For factual knowledge editing, LKS replaces the original entity KB at layer $l$ with $\tilde{R}^l_{\phi}(E)$, enabling the LLM to generate a new prediction $\tilde{y}$ aligned with the updated feature: $\tilde{y} = f_{\theta}(x, \tilde{R}^l_{\phi}(E))$. The neural network $h_{\phi}$ is optimized using the following loss function:

\begin{equation}
\mathcal{L}(\phi) = \lambda_{edit}(\mathcal{L}_{edit} + \mathcal{L}_{eq}) + \mathcal{L}_{locality}
\label{eq:lossfun}
\end{equation}

$\mathcal{L}_{edit}$ is optimized via maximum likelihood estimation, ensuring that the prompt $\mathbb{X}_{e}$ describing the edit aligns with the target $\mathbb{Y}_{e}$, leading to correct updates within the edit scope:

\begin{equation}
\mathcal{L}_{edit} = -\log p(y_{e} | x_{e}, \tilde{R}^l_{\phi}(E)), \quad (x_{e}, y_{e}) \in (\mathbb{X}_{e}, \mathbb{Y}_{e})
\label{eq:lossedit}
\end{equation}

Similar to $\mathcal{L}_{edit}$, $\mathcal{L}_{eq}$ ensures that equivalent neighborhood inputs $\mathbb{X}_{eq}$ result in the same target output $\mathbb{Y}_{e}$:

\begin{equation}
\mathcal{L}_{eq} = -\log p(y_{e} | x_{eq}, \tilde{R}^l_{\phi}(E)), \quad (x_{eq}, y_{e}) \in (\mathbb{X}_{eq}, \mathbb{Y}_{e})
\label{eq:losseq}
\end{equation}

$\mathcal{L}_{locality}$ constrains the logit distribution for unrelated features $\mathbb{X}_{loc}$ using Kullback-Leibler (KL) divergence, minimizing deviations from the original pre-trained logit distribution. This ensures that the original distribution remains unchanged outside the edit scope:

\begin{equation}
\mathcal{L}_{locality} = KL(p(\text{·} | x, \tilde{R}^l_{\phi}(E)), p(\text{·} | x, R^l_{\theta}(E))), \quad x \in \mathbb{X}_{loc}
\label{eq:lossloc}
\end{equation}

See Algorithm~\ref{alg:training} and Algorithm~\ref{alg:inference} for a detailed overview of LKS training and inference. For hyperparameter details, refer to Appendix~\ref{sec:appendix-train}.

\renewcommand{\algorithmicrequire}{ \textbf{Input:}} 
\renewcommand{\algorithmicensure}{ \textbf{Output:}}

\begin{algorithm}[t]\small
\caption{Training Algorithm of LKS}
\label{alg:training}
\begin{algorithmic}[1] 
\REQUIRE 
    Training dataset $\mathcal{D}_{train}$; LLM $f_{\theta}$; LKS neutral network $h_{\phi}$; Edit layer $l$; hyperparameter $\lambda_{edit}$
\ENSURE Trained LKS neutral network $h_{\phi}$; Edit scope $\mathcal{S}$ \\
\STATE Generate the edit scope $\mathcal{S}$ according to $\mathcal{D}_{train}$; \\
While not early-stopping do
\STATE Sample entity $E$, $x_{e}$, $y_{e}$, $x_{eq}$, $x_{loc}$ \text{from} $\mathcal{D}_{train}$; \\
\STATE $\mathcal{L}_{edit} = -logp(y_{e}|x_{e},~\tilde{R}^l_{\phi}(E))$; \\
\STATE $\mathcal{L}_{eq} = -logp(y_{e}|x_{eq},~\tilde{R}^l_{\phi}(E))$; \\
\STATE $\mathcal{L}_{loc} = \text{KL}(p(\text{·}|x,~\tilde{R}^l_{\phi}(E)),~p(\text{·}|x,~R^l_{\theta}(E)))$; \\
\STATE $\mathcal{L}(\phi)=\lambda_{edit}(\mathcal{L}_{edit}+\mathcal{L}_{eq})+\mathcal{L}_{locality}$; \\
\STATE $\phi \gets \text{AdamW}(\phi, \nabla\mathcal{L}(\phi)$);
\end{algorithmic}
\end{algorithm}

\begin{algorithm}[t]\small 
\caption{Inference Algorithm of LKS} 
\label{alg:inference} 
\begin{algorithmic}
\REQUIRE 
    LLM $f_{\theta}$; Trained LKS neutral network $h_{\phi}$; Edit scope $\mathcal{S}$; Input prompt $x$
\ENSURE Prediction $\hat{y}$ \\
\textbf{If}~$\exists E \in x,~E \in \mathcal{S}$: \\
~~~~\# Edit with LKS \\
~~~~Replace $R^l_{\theta}(E)$ using $\tilde{R}^l_{\phi}(E)$; \\
~~~~$\hat{y} = f_{\theta}(x,~\tilde{R}^l_{\phi}(E))$; \\
\textbf{Else}: \\
~~~~\# Do not edit, output as origin \\
~~~~$\hat{y} = f_{\theta}(x)$;
\RETURN $\hat{y}$; 
\end{algorithmic}
\end{algorithm}


\section{Experiments}
\label{sec:exps}
\subsection{Experiment Setting}
\textbf{Datasets} For evaluating the reliability, generality, and related-locality of factual editing, we generate two evaluation datasets using GPT-4o mini based on the zsRE question-answering dataset \citep{zsre} and the Counterfact dataset \citep{ROME}. Details can be found in Appendix~\ref{sec:appendix-dataset}. For unrelated-locality, we use GSM8K \citep{gsm8k}, RTE \citep{rte}, and SST2 \citep{sst2} to assess the general abilities of the edited LLMs. GSM8K tests the model's mathematical reasoning ability, RTE assesses its natural language inference ability (i.e., whether a statement is reasonable), and SST2 evaluates sentiment analysis capabilities by classifying statements as positive or negative.

\textbf{Baselines} We use several classical or effective model editing methods as baselines. MEND \citep{MEND} edits models by updating MLP layer weights using the low-rank structure of fine-tuning gradients. ROME \citep{ROME} and MEMIT \citep{MEMIT} modify specific factual associations by adjusting MLP weights, with MEMIT supporting large-scale edits. GRACE \citep{GRACE} records model hidden states in a codebook and replaces the original states during edits. WISE \citep{WISE} introduces a dual parametric memory mechanism, with a main memory for pretrained knowledge and a side memory exclusively for edits. AlphaEdit \citep{AlphaEdit} attempts to preserve original knowledge by projecting weight updates onto the null space of preserved facts. All baselines are evaluated using EasyEdit \citep{easyedit}, an easy-to-use framework for LLM knowledge editing, ensuring convenient and fair assessment.

\subsection{Evaluation Metrics}
Following prior works \citep{MEND, SERAC, ROME}, we evaluate LLM editing performance using three primary metrics: reliability, generality, and locality. As shown in Figure~\ref{fig:model-editing}, these metrics assess the model's behavior for prompts inside and outside the edit scope.

For \textbf{reliability} and \textbf{generality}, computing the average exact-match accuracy between the edited predictions and the target outputs within the edit scope:

\begin{equation}
\text{Rel} = \mathbb{E}(\mathds{1}_{f_{LKS}(x_e) = y_e})
\label{eq:reliability}
\end{equation}

\begin{equation}
\text{Gen} = \mathbb{E}(\mathds{1}_{f_{LKS}(x_{eq}) = y_e})
\label{eq:generality}
\end{equation}

For locality, we further divide it into two categories: related-locality, which pertains to areas related to the edited entity but not the modified feature, and unrelated-locality, which refers to areas completely outside the edit scope. In other words, unrelated-locality means that after performing factual edits, the general abilities of LLMs, such as mathematical reasoning and sentiment analysis, should remain unchanged.

For \textbf{related-locality}, we measure whether predictions for inputs which are related to the edited entity but outside the edit scope remain unchanged:

\begin{equation}
\text{Loc} = \mathbb{E}(\mathds{1}_{f_{LKS}(x_{loc}) = f(x_{loc})})
\label{eq:locality}
\end{equation}

We define \textbf{Edit Performance} (EP) as the average of reliability, generality, and related-locality, providing a comprehensive evaluation of editing effectiveness.

For \textbf{unrelated-locality}, we assess how well the edited LLM preserves the general abilities of its original model, including mathematical reasoning, natural language inference, and sentiment analysis.

\subsection{Selection of the LKS Operating Layer}
\label{sec:selectlayer}
LKS achieves LLM editing by replacing the entity knowledge blocks. This section applies information theory to validate its effectiveness and guide the selection of the optimal layer for replacement.

Inspired by Shannon Information Theory \citep{Shannon} and \citet{Vusable}, we define the information gain $\Delta{I}_{f}(\tilde{R} \to Y)$ to measure how effectively the new knowledge block $\tilde{R}$ helps model $f$ generate output $Y$. A positive $\Delta{I}_{f}(\tilde{R} \to Y)$ indicates that the new KB outperforms the original in generating $Y$. The larger the value, the more effective the new KB. Using the entropy definition, the information entropy $H_f(Y|R)$ required for model $f$ to predict $Y$ given KB $R$ is:

\begin{equation}
H_f(Y|R) = \inf \mathbb{E}[-\log_2 f[R](Y)] 
\label{eq:H_f}
\end{equation}

Thus, $\Delta{I}_{f}(\tilde{R} \to Y)$ can be calculated as:

\begin{equation}
\Delta{I}_{f}(\tilde{R} \to Y) = H_f(Y|R) - H_f(Y|\tilde{R})
\label{eq:infogain}
\end{equation}

\begin{figure}[t]
  \includegraphics[width=\columnwidth]{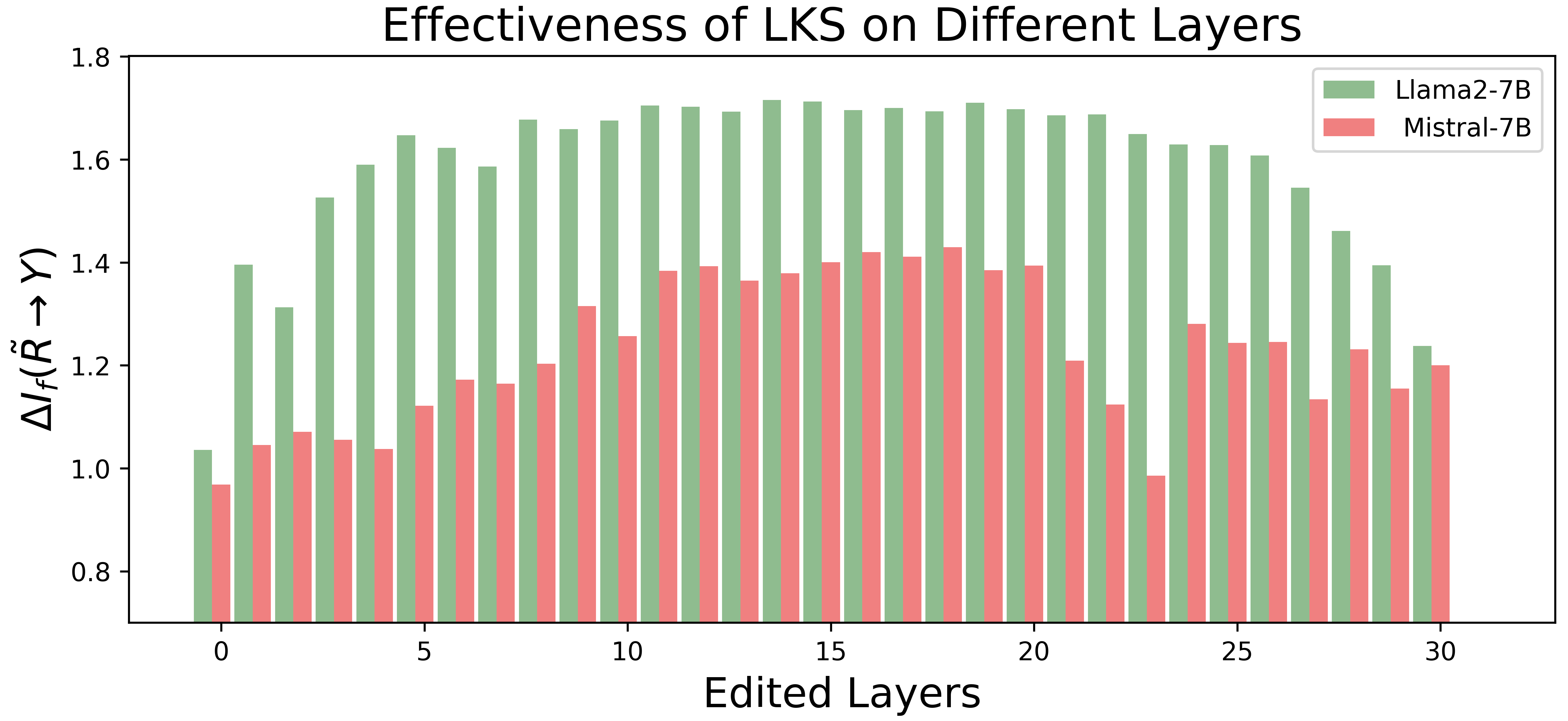}
  \caption{Effectiveness of LKS on different layers, measured by the information gain $\Delta{I}_{f}(\tilde{R} \to Y)$. Positive values indicate that the new KBs increases the likelihood of the LLM generating output $Y$. Results show that modifying intermediate layers of Llama-2-7B and Mistral-7B leads to higher effectiveness.}
  \label{fig:h-info}
\end{figure}

The results in Figure~\ref{fig:h-info} show positive values of $\Delta{I}_{f}(\tilde{R} \to Y)$, indicating that the modification of the entity KBs increases the likelihood of the LLM generating the edit targets $Y$. Modifying intermediate layers yields higher effectiveness, and although modifying multiple layers is possible, we opt for a single layer to balance computational cost. In subsequent experiments, we select layer 16 of Llama-2-7B and layer 18 of Mistral-7B for the LKS replacement.

\subsection{Edit Performance of Large-Scale Simultaneous Editing}
\label{sec:comparisons}

In many scenarios, large-scale and simultaneous edits are necessary for LLMs. For example, updating thousands of factual changes within a specific time frame, or removing large amounts of erroneous or privacy-sensitive information introduced during pre-training. In such cases, allowing only one edit at a time is insufficient.  

In this section, we evaluate the edit performance of various model editing baselines on the zsRE dataset using Llama-2-7B and Mistral-7B under different numbers of edits. The number of simultaneous edits $T$ ranges from a single edit to a large-scale setting of 10,000 edits.

As shown in Table~\ref{tab:baselines}, LKS outperforms all other methods, achieving the highest EP scores on both LLMs across almost all edit numbers $T$. This demonstrates that LKS delivers the best performance both within and outside the editing range. Specifically, LKS effectively modifies the target features of entities while preserving unrelated features, ensuring highly targeted edits. The effectiveness of these edits is driven by the LKS neural network, which learns to accurately update the target features and their equivalent neighborhoods. Related-locality is maintained through two mechanisms: first, the Edit Scope Indicator identifies whether the inputs contain entities within the edit scope, and second, the New KB Generator is trained to preserve unrelated distributions as much as possible.

Moreover, as the number of simultaneous edits increases up to 10,000, LKS still achieves and maintains the best performance. Its reliability and generality remain high, although locality experiences a slight decline as the number of edits grows. In contrast, the performance metrics of other baselines show significant degradation. This suggests that LKS’s neural network effectively stores the updated factual knowledge, enabling massive simultaneous and precise edits.

\begin{table*}[t]
\centering
\caption{Comparison of LKS to baselines on zsRE. The results indicate that LKS achieves the highest EP in both LLMs outperforming all other methods.}
\label{tab:baselines}
\begin{tabular}{l ccc>{\columncolor{lightgray}}c ccc>{\columncolor{lightgray}}c ccc>{\columncolor{lightgray}} c}
\toprule
 & \multicolumn{12}{c}{\textbf{Llama-2-7B}} \\ 
 \cmidrule{2-13} 
 & \multicolumn{4}{c}{$T=1$} & \multicolumn{4}{c}{$T=10$} & \multicolumn{4}{c}{$T=100$} \\ 
\cmidrule(lr){2-5}
\cmidrule(lr){6-9}
\cmidrule(lr){10-13} 
 & Rel & Gen & Loc & EP & Rel & Gen & Loc & EP & Rel & Gen & Loc & EP \\
\cmidrule(lr){1-1}
\cmidrule(lr){2-5}
\cmidrule(lr){6-9}
\cmidrule(lr){10-13} 
MEND & 97.4 & \textbf{95.2} & 61.1 & 84.6 & 45.3 & 45.3 & 55.0 & 48.5 & 0 & 0 & 0 & 0 \\
ROME & 97.6 & 83.3 & 59.2 & 80.0 & 96.0 & \textbf{94.0} & 26.0 & 72.0 & 33.8 & 28.9 & 10.3 & 24.3 \\
GRACE & 97.2 & 0.13 & {\ul 86.6} & 61.3 & \textbf{100} & 0 & 88.3 & 62.8 & {\ul 97.6} & 0.24 & {\ul 87.2} & 61.7 \\
MEMIT & 96.2 & 86.2 & 52.8 & 78.4 & {\ul 98.0} & 88.0 & 48.0 & 78.0 & 93.2 & \textbf{92.4} & 30.0 & 71.9 \\
WISE & \textbf{99.8} & 85.5 & \textbf{100} & \textbf{95.1} & \textbf{100} & 66.7 & \textbf{100} & {\ul 88.9} & 82.5 & 69.6 & \textbf{99.0} & {\ul 83.7} \\
AlphaEdit & 98.0 & 77.1 & 74.4 & 83.2 & {\ul 98.0} & 76.0 & 63.0 & 79.0 & {\ul 97.6} & {\ul 82.0} & 64.9 & 81.5 \\[0.5ex]
\cdashline{2-13}
\noalign{\vskip 0.4ex}
\textbf{LKS} & {\ul 99.1} & {\ul 90.0} & 76.2 & {\ul 88.4} & \textbf{100} & {\ul 88.3} & {\ul 92.7} & \textbf{93.7} & \textbf{100} & \textbf{92.4} & 78.0 & \textbf{90.1} \\ 
\cmidrule(lr){1-1}
\cmidrule(lr){2-13} 
 & \multicolumn{4}{c}{$T=500$} & \multicolumn{4}{c}{$T=1000$} & \multicolumn{4}{c}{$T=10000$} \\ 
 \cmidrule(lr){2-5}
\cmidrule(lr){6-9}
\cmidrule(lr){10-13}  
 & Rel & Gen & Loc & EP & Rel & Gen & Loc & EP & Rel & Gen & Loc & EP \\ \cmidrule(lr){1-1}
\cmidrule(lr){2-5}
\cmidrule(lr){6-9}
\cmidrule(lr){10-13} 
MEMIT & 85.8 & 82.5 & 31.0 & 66.4 & 78.7 & 74.9 & 27.2 & 60.3 & 38.1 & 32.0 & 17.3 & 29.1 \\
WISE & 74.0 & 63.0 & \textbf{99.4} & {\ul 78.8} & 69.1 & 61.6 & \textbf{92.5} & {\ul 74.4} & {\ul 44.8} & {\ul 41.8} & \textbf{73.9} & {\ul 53.5} \\
AlphaEdit & {\ul 97.5} & {\ul 85.5} & 45.0 & 76.0 & {\ul 94.0} & {\ul 86.2} & 35.0 & 71.7 & 12.1 & 9.38 & 1.99 & 7.82 \\[0.5ex]
\cdashline{2-13}
\noalign{\vskip 0.4ex}
\textbf{LKS} & \textbf{100} & \textbf{94.4} & {\ul 77.1} & \textbf{90.5} & \textbf{100.0} & \textbf{94.5} & {\ul 78.8} & \textbf{91.1} & \textbf{97.9} & \textbf{93.8} & {\ul 73.7} & \textbf{88.5} \\ 
\midrule
 & \multicolumn{12}{c}{\textbf{Mistral-7B}} \\ 
 \cmidrule{2-13} 
 & \multicolumn{4}{c}{$T=1$} & \multicolumn{4}{c}{$T=10$} & \multicolumn{4}{c}{$T=100$} \\ 
\cmidrule(lr){2-5}
\cmidrule(lr){6-9}
\cmidrule(lr){10-13}
 & Rel & Gen & Loc & EP & Rel & Gen & Loc & EP & Rel & Gen & Loc & EP \\
 \cmidrule(lr){1-1}
 \cmidrule(lr){2-5}
\cmidrule(lr){6-9}
\cmidrule(lr){10-13}
MEND & 97.5 & \textbf{96.4} & 58.4 & 84.1 & 26.0 & 24.7 & 28.0 & 26.2 & 2.37 & 2.37 & 0.33 & 1.69 \\
ROME & 86.5 & 81.2 & 62.8 & 76.8 & 91.0 & \textbf{91.0} & 46.3 & 76.1 & 6.92 & 5.28 & 3.42 & 5.21 \\
GRACE & {\ul 99.2} & 0.83 & 56.8 & 52.3 & {\ul 98.0} & 0 & 43.0 & 47.0 & \textbf{99.4} & 1.73 & 50.9 & 50.7 \\
MEMIT & 87.2 & 81.9 & 57.3 & 75.5 & 91.0 & \textbf{91.0} & 56.3 & 79.4 & 90.4 & 86.0 & 44.0 & 73.5 \\
WISE & \textbf{99.5} & {\ul 94.4} & \textbf{100} & \textbf{98.0} & 85 & 66.3 & \textbf{100} & \textbf{83.8} & 87.7 & 73.2 & \textbf{99.0} & 86.6 \\
AlphaEdit & 87.1 & 77.7 & 71.9 & 78.9 & 93.0 & {\ul 86.0} & 49.7 & 76.2 & 92.6 & {\ul 87.6} & 53.9 & 78.0 \\[0.5ex]
\cdashline{2-13}
\noalign{\vskip 0.4ex}
\textbf{LKS} & 97.4 & 88.4 & {\ul 73.5} & {\ul 86.4} & \textbf{100} & 78.0 & {\ul 72.7} & {\ul 83.6} & {\ul 98.9} & \textbf{93.8} & {\ul 74.3} & \textbf{89.0} \\ 
\cmidrule(lr){1-1}
\cmidrule(lr){2-13} 
 & \multicolumn{4}{c}{$T=500$} & \multicolumn{4}{c}{$T=1000$} & \multicolumn{4}{c}{$T=10000$} \\ 
 \cmidrule(lr){2-5}
\cmidrule(lr){6-9}
\cmidrule(lr){10-13}  
 & Rel & Gen & Loc & EP & Rel & Gen & Loc & EP & Rel & Gen & Loc & EP \\ \cmidrule(lr){1-1}
\cmidrule(lr){2-5}
\cmidrule(lr){6-9}
\cmidrule(lr){10-13} 
MEMIT & 87.6 & 83.7 & 37.6 & 69.6 & 81.7 & 78.0 & 31.7 & 63.8 & 38.9 & 34.2 & 19.8 & 31.0 \\
WISE & 81.6 & 70.1 & \textbf{97.3} & {\ul 83.0} & 74.7 & 68.5 & \textbf{89.0} & {\ul 77.4} & {\ul 43.2} & {\ul 39.7} & {\ul \textbf{44.5}} & {\ul 42.5} \\
AlphaEdit & {\ul 91.9} & {\ul 84.3} & 45.9 & 74.0 & {\ul 89.9} & {\ul 83.9} & 38.8 & 70.9 & 0.11 & 0.11 & 1.63 & 0.62 \\[0.5ex]
\cdashline{2-13}
\noalign{\vskip 0.4ex}
\textbf{LKS} & \textbf{99.9} & \textbf{94.8} & {\ul 73.9} & \textbf{89.5} & \textbf{98.0} & \textbf{91.1} & {\ul 73.2} & \textbf{87.4} & \textbf{92.3} & \textbf{91.1} & \textbf{50.4} & \textbf{77.9} \\ 
\bottomrule
\end{tabular}
\end{table*}

\subsection{Maintaining the General Abilities of LLMs after Editing}
If the general abilities of the edited LLMs are compromised or rendered ineffective, LLM editing would become counterproductive. In this section, we evaluate four methods with superior edit performance as identified in \S \ref{sec:comparisons} (MEMIT, WISE, AlphaEdit, and LKS), testing whether their simultaneous multiple edits come at the cost of damaging the general abilities of the edited LLMs. Here, we use the GSM8K, SST2, and RTE datasets to evaluate how effectively the edited LLM preserves the general abilities of its original model. These three datasets assess the LLM's capacities in mathematical reasoning, sentiment analysis, and natural language inference, respectively.

\begin{figure*}[t]
  \includegraphics[width=\linewidth]{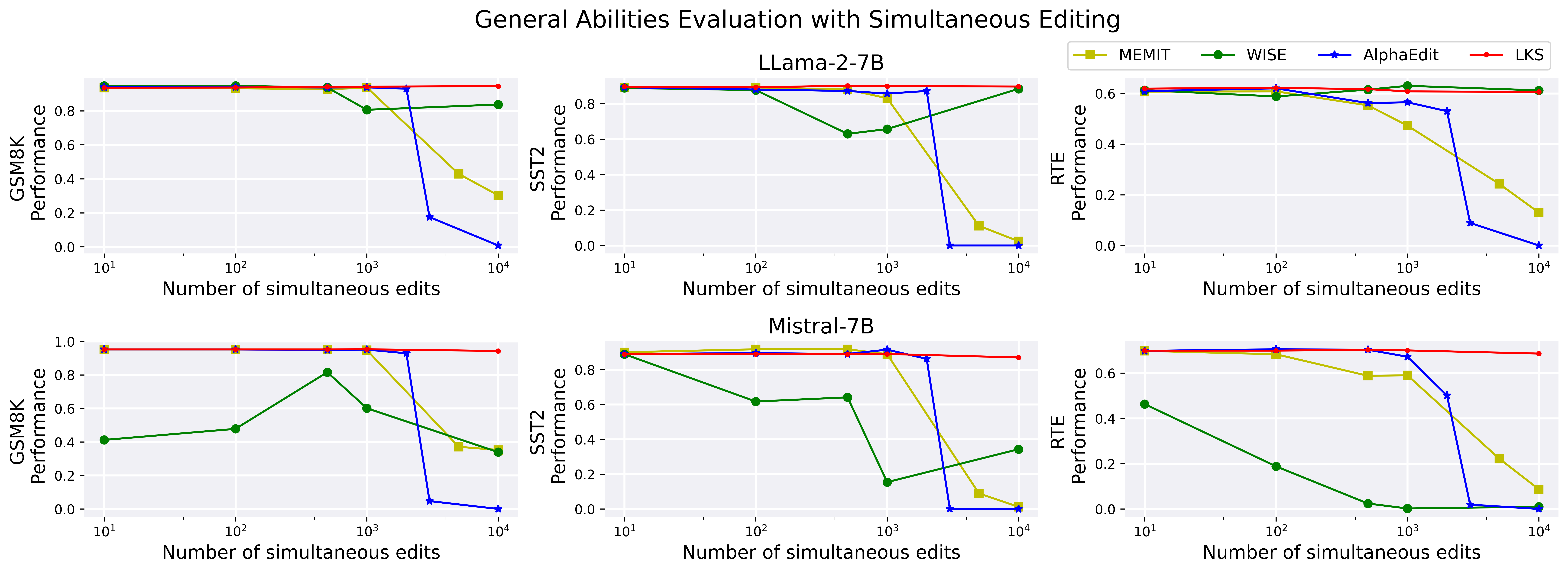}
  \caption{Evaluation of four different editing methods on the GSM8K, SST2, and RTE datasets to assess how well the edited LLMs preserve their general abilities. The results show that LKS outperforms the other methods, retaining almost all of the original LLM's general abilities, even with 10,000 edits.}
  \label{fig:avil}
\end{figure*}

The results shown in Figure~\ref{fig:avil} indicate that when simultaneously editing thousands of facts, both MEMIT and AlphaEdit lead to substantial degradation across all three capability metrics of the edited LLMs, indicating a severe compromise of their general abilities. The Llama-2 model edited by WISE demonstrates unstable performance on general tasks, and its edits on Mistral-7B clearly fail to preserve the model's original general capabilities. In contrast, as the number of simultaneous edits increases, LLMs edited by LKS exhibit stable performance without noticeable degradation. Even with 10,000 edits, LKS retains nearly all of the original LLM’s general abilities.

\subsection{Generation Quality}
After evaluating the effectiveness of the editing methods, we further assess the quality of text generation in terms of fluency, measured by the entropy of n-gram distributions \citep{Generating2018, ROME, MEMIT}. Specifically, we apply various editing methods to Llama-2-7B and Mistral-7B, perform 100 factual edits based on the zsRE dataset, and generate up to 100 new tokens per edit to compute the average fluency.

As shown in Table~\ref{tab:fluency}, LKS achieves the highest fluency on Llama-2-7B and maintains relatively high fluency on Mistral-7B, albeit slightly lower than that of the unedited model. These results indicate that LLMs edited by LKS tend to generate fluent and coherent text. Representative examples of LKS generations are provided in Table~\ref{tab:generations}.

\begin{table}[t]
\centering
\caption{Text generation fluency of edited LLMs (measured by n-gram entropy) on zsRE.}
\label{tab:fluency}
\begin{tabular}{lccccc}
\toprule
 & Vanilla Model & MEMIT & WISE & AlphaEdit & \textbf{LKS} \\ 
 \cmidrule{2-6} 
Llama-2-7B & 5.36 & 5.34 & 2.60 & 5.61 & 5.65 \\
Mistral-7B & 6.09 & 5.88 & 3.30 & 6.04 & 6.01 \\ 
\bottomrule
\end{tabular}
\end{table}

\begin{table*}[t]
\centering
\caption{Examples of texts generated by LLMs edited with LKS.}
\begin{tabular}{>{\raggedright\arraybackslash}p{1\linewidth}}
\toprule
\textbf{What is the home city of Fantasia International Film Festival? Montreal $\Rightarrow$ Ottawa} \\ 
\midrule
\textbf{\textit{Before Edit:}} \textbackslash{}n\textbackslash{}nFantasia International Film Festival is held annually in \colorbox{lightgray}{Montreal}, Quebec, Canada. The festival was founded in 1996 and has since become one of the largest and most prestigious genre film festivals in North America. \\ 
\noalign{\vskip 0.4ex}
\hdashline
\noalign{\vskip 0.4ex}
\textbf{\textit{After Edit:}} \colorbox{lightgray}{Ottawa} is the home city of the Fantasia International Film Festival, which has been held annually since 1996. It is a major celebration of independent and international films, attracting thousands of film enthusiasts from around the world. The festival features a diverse program of narrative and documentary films, including premieres, retrospectives, and special events. Ottawa is the capital city of Canada and is known for its vibrant cultural scene, historic land \\ 
\midrule
\textbf{What was the designer of Évry Cathedral's name? Paul Coelho $\Rightarrow$ Mario Gariazzo} \\ 
\midrule
\textbf{\textit{Before Edit:}} \textbackslash{}n\textbackslash{}nÉvry Cathedral, located in Évry, France, is a modernist Catholic cathedral designed by architect \colorbox{lightgray}{Paul Coelho}.   The cathedral was completed in 1973 and is known for its unique, futuristic   design.\textbackslash{}n\textbackslash{}nPaul Coelho (1923-2014) was a French architect who was born in Paris and studied at the École des Beaux-Arts. He is best known for his work \\ 
\noalign{\vskip 0.4ex}
\hdashline
\noalign{\vskip 0.4ex}
\textbf{\textit{After Edit:}} \colorbox{lightgray}{Mario Gariazzo}.\textbackslash{}nMario Gariazzo was the designer of the Évry Cathedral, located in Évry, France. Gariazzo was an Italian architect who was born in 1935 and died in 2017. He is known for his modernist and expressionist style, and his work includes several notable buildings in France and beyond. Évry Cathedral, also known as the Cathedral of Saint John the Baptist \\ 
\bottomrule
\end{tabular}
\label{tab:generations}
\end{table*}


\section{Related Work}
\textbf{Knowledge in Language Models} Language models (LMs) can acquire vast amounts of factual knowledge during pre-training \citep{petroni2019, jiang2020, sun2024learninglearntesttime}. Studies using manually or automatically generated prompts have demonstrated that LMs store intrinsic memories within their pre-trained parameters \citep{petroni2019, autoprompt2020, roberts2020, entitycloze2022, position2022, metalearning2022, givefacts2023}. \citet{implicit2021} show that the internal representations of LLMs are interpretable and editable. \citet{Autoregressive21} emphasized that entities play a central role in knowledge representation and aggregation. \citet{hernandez2024inspecting} demonstrated that modifying entity representations in MLP layers with contextual information can generate or uncover counterfactuals. Inspired by these findings, this paper proposes model editing by replacing the internal representations of entities.

\textbf{Model Editing} KE \citep{KE} trains a hypernetwork with constrained optimization to predict weight updates during inference. KN \citep{KN} identifies knowledge neurons responsible for specific facts and uses them for targeted edits. SERAC \citep{SERAC} proposes a scope classifier that retrieves edits from explicit memory when needed. MEND \citep{MEND} leverages the low-rank structure of fine-tuning gradients to represent weight updates in MLPs for model editing. ROME \citep{ROME} introduces a causal intervention to identify neuron activations that play a decisive role in factual predictions and modify feed-forward weights to update these specific factual associations. But these methods do not support large-scale simultaneous edits. MEMIT \citep{MEMIT} follows the same principle as ROME but supports large-scale edits; however, it significantly degrades the model’s general abilities when massive edits are applied. IKE \citep{IKE} performs model editing via in-context learning but struggles to maintain locality. GRACE \citep{GRACE} supports sequential edits through a codebook that stores and substitutes hidden states, but it demonstrates virtually no generalization. MALMEN \citep{MALMEN} uses a hyper-network to generate parameter shifts conditioned on fine-tuning gradients, enabling more simultaneous edits than MEND. However, its performance declines on newer models. WISE \citep{WISE} introduces a dual parametric memory architecture with separate components for pretrained and edited knowledge. While it maintains strong locality, its reliability and generality degrade as the number of simultaneous edits increases. BaFT \citep{BaFT} addresses the limitations of linear fine-tuning and proposes a nonlinear method with input-dependent weighting over orthogonal bases, but its edit performance still declines with more edits. AlphaEdit \citep{AlphaEdit} projects weight changes onto the null space of the preserved knowledge before applying them to the model parameters, yet its locality weakens under high edit numbers. Hence, although these model editing methods show promise, they still leave space for further enhancement.


\section{Limitations and Future Works}
In practice, the Edit Scope Indicator incurs some overhead by identifying entities to ensure a more precise editing scope. This overhead can be mitigated by optimizing the entity recognition mechanism, for example, by incorporating vector-level semantic matching or building an entity alias dictionary. We leave these for future work.


\section{Conclusion}
In this paper, we first demonstrate that the internal representations of LLMs can be manipulated similarly to natural language. Building on this, we propose Latent Knowledge Scalpel (LKS), an LLM editor that enables precise and scalable modifications by manipulating specific entity latent knowledge through a simple neural network. Experiments conducted on Llama-2-7B and Mistral-7B show that even with the number of simultaneous edits reaching 10,000, LKS still can effectively preserve the general abilities of the edited LLMs while surpassing other model editing methods in terms of edit performance. Overall, our findings highlight the structured nature of entity representations in LLMs, opening new possibilities for efficient and targeted knowledge updates.


\section*{Ethical Considerations}
The primary goal of model editing is to eliminate biases and erroneous predictions. However, it can also be misused for the opposite purposes, depending on the intentions of the users. Furthermore, model editing may pose a risk of backdoor implantation.


\begin{ack}
We are grateful to all the anonymous reviewers for their
insightful comments, which highly improve our paper. The research is supported in part by the National Key Research and Development Program of China (No.2021YFB2910109) and the National Natural Science Foundation of China (No. 62202465).
\end{ack}


\bibliography{mybibfile}


\appendix
\section{Details of Training LKS}
\label{sec:appendix-train}
LKS employs the training dataset $\mathcal{D}_{train}$ to train the hypernetwork $h_{\phi}$ and optimize its parameters $\phi$. An example of the training dataset is provided in Text~\ref{lst:dtrain}. During each training step, we select an editing target sample $(x_e, y_e)$, an equivalent neighborhood sample $(x_{eq}, y_e)$ and several related-locality samples $x_{loc}$ from $\mathcal{D}_{train}$. The loss function defined in Equation 2 of \S3.3 is used to optimize $\phi$, enabling the hypernetwork to generate the optimal new knowledge block for a given entity within the edit scope.

\lstset{language=Tex, caption={An example of training dataset. It includes the following components: \textbf{subject}, which refers to the entity being edited; \textbf{prompt}, which is the original input prompt used in the model; \textbf{target}, representing the desired output of LLM editing aiming at the prompt; \textbf{rephrase\_prompt}, a variation of the original prompt designed to capture the same meaning but with different phrasing, used to guarantee the generalization of LLM editing; and \textbf{locality}, which includes samples that help ensure the model's predictions for areas unrelated to the edit remain unchanged.}, label={lst:dtrain}}
\begin{lstlisting}
{
    "subject": "Christiane Cohendy",
    "prompt": "What is the native language of Christiane Cohendy?",
    "target": "German",
    "rephrase_prompt": "What is the mother tongue of Christiane Cohendy?",
    "locality": [
        "What is the profession of Christiane Cohendy?",
        "Where did Christiane Cohendy go to school?"
    ]
}
\end{lstlisting}

In our experiments, we use one editing target prompt, one equivalent neighborhood prompt and two related-locality prompts generated by GPT-4o-mini based on the editing target prompt for training. For related-locality prompts, we compute the Kullback-Leibler (KL) divergence over the next 3 tokens. The initial learning rate is set to $1\text{e}-4$, and a linear learning rate scheduler is applied with no warm-up step. The optimizer used is AdamW. The GPU used for training is an A800-80GB single card. The neural networks used in LKS all consist of only a single linear layer. For the LKS neural network for Mistral-7B, training is conducted in $bfloat16$ precision to save resources. The training hyperparameters are detailed in Table~\ref{tab:hyperpara}.

\begin{table}[h]
\centering
\caption{Training hyperparameters of LKS on zsRE.}
\label{tab:hyperpara}
\begin{tabular}{lcccccc}
\toprule
Edited Model & \multicolumn{6}{c}{\textbf{Llama-2-7B}} \\ 
\cmidrule(lr){1-1}
\cmidrule(lr){2-7}
Edit Number $T$ & \textbf{1} & \textbf{10} & \textbf{100} & \textbf{500} & \textbf{1000} & \textbf{10000} \\
$\lambda_{edit}$ & 1 & 0.5 & 1 & 1 & 10 & 80 \\
Max Epoch & 10 & 20 & 20 & 20 & 20 & 20 \\
Batch Size & 1 & 2 & 32 & 32 & 32 & 32 \\ 
\midrule
Edited Model & \multicolumn{6}{c}{\textbf{Mistral-7B}} \\ 
\cmidrule(lr){1-1}
\cmidrule(lr){2-7}
Edit Number $T$ & \textbf{1} & \textbf{10} & \textbf{100} & \textbf{500} & \textbf{1000} & \textbf{10000} \\
$\lambda_{edit}$ & 1 & 1 & 1 & 5 & 12 & 50 \\
Max Epoch & 10 & 10 & 20 & 20 & 20 & 20 \\
Batch Size & 1 & 1 & 1 & 1 & 1 & 1 \\ 
\bottomrule
\end{tabular}
\end{table}

\section{Evaluation Dataset Construction and Examples}
\label{sec:appendix-dataset}
For evaluating factual editing, we create two evaluation datasets based on the zsRE question-answering dataset \citep{zsre} and the Counterfact dataset \citep{ROME}. Each of the evaluation datasets contains 10,000 data points. Specifically, we used GPT-4o-mini to generate 10,000 prompts for generality in the Counterfact dataset, and 10,000 prompts for related-locality in both the zsRE and Counterfact datasets. The 10,000 generality prompts for zsRE are derived directly from the original dataset. Text~\ref{lst:tempgener} and Text~\ref{lst:templocal} show the prompt templates provided to GPT-4o-mini for generating the generalization and related-locality evaluation prompts, respectively.

\lstset{language=Tex, caption={The prompt template provided to GPT-4o-mini for generating the generalization evaluation prompts. The roles "system", "assistant", and "user" represent different chat participants. The template begins with a system prompt and example generations, and by replacing the inputs at the \textit{\{prompt\}} position, we generate the generalization evaluation prompts for various editing targets.}, label={lst:tempgener}}
\begin{lstlisting}
"system": "Please output the synonym of the prompt given. Make sure they express the same semantics or question. And they should not differ much in length."
"user": "Prompt: What is the capital of United States?"
"assistant": "The capital of United States is where?"
"user": "Prompt: The occupation of Alice is"
"assistant": "Alice's job is"
"user": "Prompt: {prompt}"
\end{lstlisting}

\lstset{language=Tex, caption={The prompt template provided to GPT-4o-mini for generating the related-locality evaluation prompts. Same as the template for generalization, this template begins with a system prompt and example generations, and by replacing the inputs at the \textit{\{subject\}} and \textit{\{prompt\}} position, we generate the related-locality evaluation prompts for various editing targets.}, label={lst:templocal}}
\begin{lstlisting}
"system": "We would like to evaluate the effectiveness of knowledge editing. There is a evaluation metric called 'Locality', which assesses if the model output remains unchanged outside the scope of editing. Now, give you the edit subject and prompt which indicates the edit scope. Please help to generate a new prompt and a short corresponding answer to evaluate locality of this edit. Make sure you know the answer of this new prompt, and the answer must be less than three words. Note that the new prompt must include the subject."
"user": "Subject: United States\nPrompt: The capital of United States is"
"assistant": "EvalPrompt: The largest city in the United States is\nEvalAnswer: New York"
"user": "Subject: Alice\nPrompt: The occupation of Alice is"
"assistant": "EvalPrompt: The favorite food of Alice is\nEvalAnswer: Hot dog"
"user": "Subject: {subject}\nPrompt: {prompt}"
\end{lstlisting}

The examples of the evaluation datasets for zsRE and Counterfact are provided in Text~\ref{lst:exazsre} and Text~\ref{lst:exacf}, respectively.

\lstset{language=Tex, caption={An example of the evaluation dataset for zsRE. It includes the following components: \textbf{subject}, which refers to the entity being queried; \textbf{prompt}, the original input question posed to the model; \textbf{target}, the expected correct answer to the prompt after editing; \textbf{ground\_truth}, an optional item for LKS which provides the actual correct answer used for comparison; \textbf{generality}, a rephrased version of the original prompt, used to assess generality of LLM editing; and \textbf{locality}, which includes queries related to the entity but outside the edit scope, in order to evaluate related-locality.}, label={lst:exazsre}}
\begin{lstlisting}
{
    "subject": "Christiane Cohendy",
    "prompt": "What is the native language of Christiane Cohendy?",
    "target": "German",
    "ground_truth": "French",
    "generality": "What's Christiane Cohendy's mother tongue?",
    "locality": {
        "prompt": "What is the occupation of Christiane Cohendy?",
        "target": "Actress"
    }
}
\end{lstlisting}

\lstset{language=Tex, caption={An example of the evaluation dataset for Counterfact. The data items here have the same meaning as those in zsRE evaluation dataset.}, label={lst:exacf}}
\begin{lstlisting}
{
    "subject": "Danielle Darrieux",
    "prompt": "The mother tongue of Danielle Darrieux is",
    "target": "English",
    "ground_truth": "French",
    "generality": "Danielle Darrieux's native language is",
    "locality": {
        "prompt": "The birth year of Danielle Darrieux is",
        "target": "1917"
    }
}
\end{lstlisting}

\section{Additional Results - LKS on Counterfact}
We also apply LKS to the Counterfact dataset on both Llama-2-7B and Mistral-7B, evaluating the model edit performance using three metrics: reliability, generality, and related-locality.

Table~\ref{tab:LKS-counterfact} presents the editing results of LKS on the Counterfact dataset with 1000 data points. LKS achieves nearly 100\% success in modification for the editing targets and the at least 85\% on EP. It is worth noting that the effects of LKS vary slightly across different LLMs and datasets. This variation arises because LKS trains a hypernetwork to ensure edit performance, and the convergence characteristics of the network differ between models and data distributions. Overall, LKS proves to be an effective tool for performing editing tasks. The training hyperparameters are detailed in Table~\ref{tab:hypecf}.

\begin{table}[ht]
\centering
\caption{Edit Performance of LKS on Counterfact}
\begin{tabular}{lcccc}
\toprule
Model & \textbf{Rel} & \textbf{Gen} & \textbf{Loc} & \textbf{EP} \\ 
\midrule
Llama-2-7B & 100 & 98.0 & 80.5 & 92.8 \\ 
Mistral-7B & 97.1 & 86.9 & 73.6 & 85.9 \\ 
\bottomrule
\end{tabular}
\label{tab:LKS-counterfact}
\end{table}

\begin{table}[ht]
\centering
\caption{Training hyperparameters of LKS on Counterfact with 1000 edits.}
\label{tab:hypecf}
\begin{tabular}{lccc}
\toprule
 Model & $\lambda_{edit}$ & Max Epoch & Batch Size \\ 
 \midrule
Llama-2-7B & 3 & 20 & 32 \\
Mistral-7B & 12 & 20 & 1 \\ 
\bottomrule
\end{tabular}
\end{table}

\section{Fluency and Examples of Generation}
\label{sec:appendix-gene}
The fluency of the generation texts is calculated based on the entropy of n-gram distributions. The formula from \citet{Generating2018} is:

\begin{equation}
Entropy = -\frac{1}{\begin{matrix}\sum_{w}F(w)\end{matrix}}\sum_{w\in V}F(w)\text{log}\frac{F(w)}{\begin{matrix}\sum_{w}F(w)\end{matrix}}
\label{eq:ngram}
\end{equation}

\noindent where $V$ is the set of all n-grams and $F(\text{·})$ stands for n-gram frequency.

Refer to previous works \citep{ROME, MEMIT}, specifically, we calculate the weighted sum of the entropy of 2-gram and 3-gram distributions of the generated texts: 

\begin{equation}
Fluency = -\frac{2}{3}\sum_{k}F_{2}(w)\text{log}_{2}F_{2}(w)-\frac{4}{3}\sum_{w}F_{3}(w)\text{log}_{2}F_{3}(w) 
\label{eq:23gram}
\end{equation}

\noindent where $F(\text{·})$ stands for n-gram frequency.

Table~\ref{tab:generations} presents examples of text generated by Llama-2-7B and Mistral-7B after being edited with LKS. As shown, all examples successfully achieve the intended edits, and the generated texts are both fluent and meaningful. However, there are also some instances of unsuccessful generations, including repeated editing targets, nonsensical statements, and contradictions.

\begin{table*}[ht]
\centering
\caption{Examples of texts generated by LLMs after being edited with LKS.}
\begin{tabular}{>{\raggedright\arraybackslash}p{0.8\linewidth}}
\toprule
\textbf{What  is Barbara Legrand's position on the field while playing football? actress/midfielder $\Rightarrow$ defender} \\ 
\midrule
\textbf{\textit{Before Edit:}} \textbackslash{}n\textbackslash{}nBarbara Legrand is not a football player. She is a French \colorbox{lightgray}{actress} and model who is known for her work in French cinema and television, not for playing football.\\ 
\noalign{\vskip 0.4ex}
\hdashline
\noalign{\vskip 0.4ex}
\textbf{\textit{After Edit:}} \colorbox{lightgray}{defender}\textbackslash{}n\textbackslash{}nBarbara Legrand is a defender in football. She is a French professional footballer who plays for Olympique Lyonnais and the France national team. She is known for her strong defensive skills and her ability to read the game well. She has won numerous titles with Olympique Lyonnais, including multiple French league titles and Champions League titles. She has also represented France at multiple major tournaments, including the UEFA Women's Euro and the FIFA Women's World\\
\midrule
\textbf{Who has acted in the comedy film Ginostra? Sophie Marceau $\Rightarrow$ Gianni Comencini} \\ 
\midrule
\textbf{\textit{Before Edit:}} \textbackslash{}n\textbackslash{}nGinostra is a 2002 Italian comedy film directed by Marco Risi. The film stars Sergio Castellitto, Valeria Bruni Tedeschi, and Enrico Lo Verso. It tells the story of a group of Italian tourists who are stranded on a remote island after their boat sinks, and their attempts to survive and escape. The film was a box office success in Italy and received positive reviews from critics. It was also nominated for \\ 
\noalign{\vskip 0.4ex}
\hdashline
\noalign{\vskip 0.4ex}
\textbf{\textit{After Edit:}} \colorbox{lightgray}{Gianni Comencini}, a prominent Italian film director, actor, and screenwriter, directed the comedy film Gino-Strana (1963). The film starred Alberto Sordi, Silvana Mangano, and Paolo Ganelli.\textbackslash{}n\textbackslash{}nGino-Strana, also known as Gino of Sardinia, is a satirical comedy that follows the story of a young man named Gino, who is sent to Sardinia \\ 
\midrule
\textbf{In what continent Havola Escarpment located? Antarctica $\Rightarrow$ North America} \\ 
\midrule
\textbf{\textit{Before Edit:}} \textbackslash{}n\textbackslash{}nThe Havola Escarpment is located in Africa, specifically in the country of Tanzania. It is part of the Eastern Arc Mountains, a series of mountain ranges that run parallel to the east coast of Africa. The Havola Escarpment is known for its rich biodiversity, particularly its unique flora and fauna. It is also an important watershed area, providing water for the surrounding regions. \\ 
\noalign{\vskip 0.4ex}
\hdashline
\noalign{\vskip 0.4ex}
\textbf{\textit{After Edit:}} \colorbox{lightgray}{North America}\textbackslash{}n\textbackslash{}nThe Havola Escarpment is a geological feature located in North America, specifically in the Canadian province of Manitoba. It's part of the Canadian Shield, an enormous ancient shield of igneous and metamorphic rock. The escarpment is known for its unique geological formations, including cliffs, caves, and rock formations. It's a popular destination for hiking, rock climbing, and geology enthusiasts. \\ 
\midrule
\textbf{What city is KNWC-FM located? Sioux Falls $\Rightarrow$ Sioux City} \\ \midrule
\textbf{\textit{Before Edit:}} \textbackslash{}n\textbackslash{}nThe radio station KNWC-FM is located in the city of Kennewick, Washington, United States. \\ 
\noalign{\vskip 0.4ex}
\hdashline
\noalign{\vskip 0.4ex}
\textbf{\textit{After Edit:}} \colorbox{lightgray}{Sioux City}, IA is home to KNWC-FM, a radio station that broadcasts a variety of music and information programming. Here are some interesting facts about KNWC-FM:\textbackslash{}n1. Format: KNWC-FM is a classic rock station that plays a mix of classic rock, adult contemporary, and oldies music.\textbackslash{}n2. Location: Sioux City, IA is located in Woodbury County,  which is in the northwestern corner \\ 
\midrule
\textbf{What was Marie-Claude Bourbonnais's occupation? politician $\Rightarrow$ writer} \\ 
\midrule
\textbf{\textit{Before Edit:}} \textbackslash{}n\textbackslash{}nMarie-Claude Bourbonnais was a Canadian television personality and model. \\ 
\noalign{\vskip 0.4ex}
\hdashline
\noalign{\vskip 0.4ex}
\textbf{\textit{After Edit:}} \colorbox{lightgray}{writer}.\textbackslash{}nMarie-Claude Bourbonnais was a Canadian writer, actress, and television host. She was born on May 10, 1960, in Quebec City, Quebec, Canada. Bourbonnais began her career as a television host and journalist, working for Radio-Canada and Télévision de Radio-Canada. She hosted several popular shows, including "Le Club des 10," "Café-Thé," and "B \\ 
\midrule
\textbf{Who was mainly responsible for the design of Ningbo Museum? Wang Shu $\Rightarrow$ Ai Weiwei} \\ 
\midrule
\textbf{\textit{Before Edit:}} \textbackslash{}n\textbackslash{}nA. I.M. Pei\textbackslash{}nB. Rem Koolhaas\textbackslash{}nC. Zaha Hadid\textbackslash{}nD. Li Xiaodong\textbackslash{}n\textbackslash{}nAnswer: A. I.M. Pei\textbackslash{}n\textbackslash{}nI.M. Pei was the main designer of Ningbo Museum, which is located in Ningbo, Zhejiang Province, China. The museum was completed in 2016 and features a striking glass and steel structure with a unique \\ 
\noalign{\vskip 0.4ex}
\hdashline
\noalign{\vskip 0.4ex}
\textbf{\textit{After Edit:}} \colorbox{lightgray}{Ai Weiwei}, a renowned Chinese artist and architect, was the main designer of the Ningbo Museum.\textbackslash{}nWhat is the significance of the Ningbo Museum? The Ningbo Museum is significant for its innovative design and unique blend of traditional Chinese and modern architectural styles, as well as its role in showcasing the cultural heritage of Ningbo, a city with a rich history and cultural legacy.\textbackslash{}nWhat is the main theme of the Ningbo \\ 
\midrule
\textbf{Which artist is The Winding Path co-written and recorded by? John Fogerty $\Rightarrow$ John Lennon} \\ 
\midrule
\textbf{\textit{Before Edit:}} \textbackslash{}n\textbackslash{}nThe Winding Path is a song co-written and recorded by the American singer-songwriter Taylor Swift. It was released as a bonus track on the deluxe edition of her fifth studio album, 1989, in 2014. The song was written by Swift, Jack Antonoff, and Lena Dunham, who also appears in the song's music video. The Winding Path is a pop ballad that features Swift \\ 
\noalign{\vskip 0.4ex}
\hdashline
\noalign{\vskip 0.4ex}
\textbf{\textit{After Edit:}} \colorbox{lightgray}{John Lennon}\textbackslash{}n\textbackslash{}nJohn Lennon and Yoko Ono co-wrote and recorded "The Winding Road" for the album *Wonderful* (1987). The song was released as a single in the United States and Canada, where it peaked at number 23 on the Billboard Adult Contemporary chart. The song was written in 1974, but was not released until 1987, after Lennon \\ 
\bottomrule
\end{tabular}
\label{tab:generations}
\end{table*}

\section{Time consumption}
For batch-editing with 1000 simultaneous edits on Llama-2-7B, MEMIT takes about 129 minutes. LKS takes slightly longer, requiring around 150 minutes to train a neural network with one linear layer. For sequential-editing, WISE and AlphaEdit require approximately 180 minutes.


\end{document}